\pdfoutput=1
\PassOptionsToPackage{table}{xcolor}

\documentclass[11pt]{article}

\usepackage[final]{acl}

\usepackage{times}
\usepackage{latexsym}

\usepackage[T1]{fontenc}
\usepackage[utf8]{inputenc}

\usepackage{microtype}

\usepackage{inconsolata}

\usepackage[pdftex]{graphicx}
\usepackage{booktabs}
\usepackage{array}
\usepackage{tikz}
\usepackage{tabularx}
\usetikzlibrary{positioning}
\usetikzlibrary{tikzmark}
\usepackage{enumitem}
\usepackage{comment}
\usepackage{amsmath}
\usepackage[most]{tcolorbox}
\usepackage{fontawesome5}
\usepackage{todonotes}

\definecolor{mylightblue}{RGB}{207, 233, 242}

\newcommand{\mycircle}[1]{{\small \textcolor{#1}{\faCircle}}}

\usepackage{etoolbox}
\usepackage{contour}
\contourlength{0.5pt}
\newrobustcmd{\contourcircle}[2]{\raisebox{0.2pt}{\scalebox{0.91}{\contour{#2}{\small \textcolor{#1}{\faCircle}}}}}

\newcommand{\suppositionstrat}{\contourcircle{green!40}{green!20}}
\newcommand{\chainstrat}{\contourcircle{yellow!40}{yellow}}
\newcommand{\compoundstrat}{\mycircle{orange!20}}
\newcommand{\concatstrat}{\mycircle{mylightblue}}
\newcommand{\symbolicstrat}{\mycircle{blue!20}}

\newcommand{\LLaMASmall}{LLaMA-2-7B}
\newcommand{\LLaMAMedium}{LLaMA-2-13B}
\newcommand{\LLaMABig}{LLaMA-2-70B}

\title{Comparing Inferential Strategies of Humans and Large Language Models in Deductive Reasoning}

\author{Philipp Mondorf\textsuperscript{\normalfont 1, 2} \and
         Barbara Plank\textsuperscript{\normalfont 1, 2}\\
  \textrm{\textsuperscript{1}}MaiNLP, Center for Information and Language Processing, LMU Munich, Germany \\
  \textrm{\textsuperscript{2}}Munich Center for Machine Learning (MCML), Munich, Germany \\
{\tt p.mondorf@lmu.de} \hspace{2em} {\tt b.plank@lmu.de}}

\begin{document}

\maketitle
\begin{abstract}
Deductive reasoning plays a pivotal role in the formulation of sound and cohesive arguments. It allows individuals to draw conclusions that logically follow, given the truth value of the information provided. Recent progress in the domain of large language models (LLMs) has showcased their capability in executing deductive reasoning tasks. Nonetheless, a significant portion of research primarily assesses the accuracy of LLMs in solving such tasks, often overlooking a deeper analysis of their reasoning behavior. In this study, we draw upon principles from cognitive psychology to examine inferential strategies employed by LLMs, through a detailed evaluation of their responses to propositional logic problems. Our findings indicate that LLMs display reasoning patterns akin to those observed in humans, including strategies like \textit{supposition following} or \textit{chain construction}. Moreover, our research demonstrates that the architecture and scale of the model significantly affect its preferred method of reasoning, with more advanced models tending to adopt strategies more frequently than less sophisticated ones. Importantly, we assert that a model's accuracy, that is the correctness of its final conclusion, does not necessarily reflect the validity of its reasoning process. This distinction underscores the necessity for more nuanced evaluation procedures in the field.
\end{abstract}

\section{Introduction}\label{sec:Introduction}

% Opening Figure
\begin{figure}[t!]
  \centering 
  \begin{tikzpicture}[
    font=\fontsize{8}{9.6}\selectfont, % Sets the font size globally for the tikzpicture
    box/.style={draw, rectangle, rounded corners, minimum height=1cm, align=left}, % Sets text alignment to left
    wide box/.style={draw, rectangle, rounded corners, minimum height=1cm, text width=7.5cm, align=left, fill=gray!5}, % Sets text alignment to left for the wide box
    green box/.style={box, fill=green!20, text width=3.55cm}, % Increased text width and modified style for the green box
    yellow box/.style={box, fill=yellow!20, text width=3.55cm} % Increased text width and modified style for the yellow box
    ]
    \node[wide box] (top) at (0,0) {
    \textbf{Task Prompt:}\\Evaluate whether the given conclusion logically follows from the premises below:

    % Customizing the enumerate list
    \begingroup
    \setlist[enumerate]{itemsep=0pt, parsep=0pt, partopsep=0pt, topsep=0pt, leftmargin=*} % Removes space between list items and adjusts list margins
    \begin{enumerate}
        \item There is a pink marble if and only if there is an olive marble.
        \item If there is an olive marble then there is a brown marble.
        \item Either there is a brown marble or else there is a black marble, but not both.
    \end{enumerate}
    \endgroup

    Conclusion: Either there is a pink marble or else there is a black marble, but not both.
    };
    \node[green box, below=0.01\textwidth of top.south west, anchor=north west] (left)  { % Adjusted position for the left box
    \textbf{Answer:}\\
    Okay, let's say there is a pink marble.\\According to the first statement, there must be an olive marble. Then, according to the second statement, there must be a brown marble. But wait, that means there cannot be a black marble, because of the third statement!\\So, the conclusion is actually true.
    };
    \node[yellow box, below=0.01\textwidth of top.south east, anchor=north east] (right) { % Adjusted position for the right box
    \textbf{Answer:}\\
    \begingroup
    \setlist[enumerate]{itemsep=0pt, parsep=0pt, partopsep=0pt, topsep=0pt, leftmargin=*} % Removes space between list items and adjusts list margins
    \begin{enumerate}
        \item If there is a pink marble, then there must be an olive marble.
        \item If there is an olive marble, then there must be a brown marble.
        \item If there is a brown marble, then there cannot be a black marble.
    \end{enumerate}
    \endgroup
    Therefore, if there is a pink marble, then there cannot be a black marble.\\Conclusion: True
    };
\end{tikzpicture}
  \caption{
  Given the propositional reasoning prompt (top box), the LLM shows two different inferential strategies: \textbf{supposition following} (left) and \textbf{chain construction} (right), see Section~\ref{sec:Strategies-Propositional-Reasoning} for strategy details. Note that both answers are only partially correct, as the exclusive disjunction has only been proven for one of the cases (pink and not black). Model responses are generated by \texttt{LLaMA-2-Chat-70B} across two random seeds.
  }
  \label{fig:model_strategies_example}
\end{figure}

% Start of Introduction
Deductive reasoning, that is the process of drawing conclusions that logically follow from the information at hand, is an integral aspect of human cognition and plays a pivotal role in formulating sound and coherent arguments \cite{leighton_defining_2003}. Take, for example, the following statements:

\vspace{1mm}

\noindent\textit{If there is a blue marble in the box then there is a green marble in the box.} \\
\noindent\textit{There is a blue marble in the box.}

\vspace{1mm}

\noindent Even without proper training in logic, most individuals can naturally deduce the valid conclusion:

\noindent\textit{Therefore, there is a green marble in the box.}

\noindent This innate capability of drawing conclusions that invariably follow from the truth value of available information has been a focal point of scholarly interest for centuries \cite{holyoak_cambridge_2005}. Propositional logic, a subfield of deductive reasoning, focuses on constructing logical arguments based on the relationship between statements similar to those in the example previously mentioned \cite{hurley_concise_2011}. Extensive research has been dedicated to examining human reasoning behavior in contexts that involve propositional logic. For instance, ~\citet{van_der_henst_strategies_2002} have identified \emph{five different strategies} people commonly employ when navigating problems of propositional logic (see Section \ref{sec:Strategies-Propositional-Reasoning}). Such behavioral studies have been crucial in shaping theories that shed light on the fundamental elements of cognitive reasoning processes \cite{rips_psychology_1994, johnson-laird_mental_1986, kahneman_judgment_1982}.

In parallel, recent advancements in the field of large language models have demonstrated their potential in executing tasks involving deductive reasoning \cite{yang_logical_2023, yu_natural_2024, huang_towards_2023}. Yet, the extent to which LLMs truly possess such abilities remains a subject of ongoing debate \cite{mahowald_dissociating_2024, mitchell_debate_2023}. Unlike behavioral studies in human reasoning that are often characterized by in-depth examinations of the reasoners' expressions, many studies on LLM-based reasoning tend to focus on task performance and accuracy metrics, offering limited insights into the underlying reasoning behavior of the models \cite{mitra_orca_2023, openai_gpt-4_2023, gemini_team_gemini_2023}.

In this paper, we draw from the cognitive science literature~\cite{van_der_henst_strategies_2002} and study inferential strategies employed by LLMs when solving propositional logic problems (see Figure \ref{fig:model_strategies_example}). We analyze the reasoning behavior of three different language model families, varying in model size and fine-tuning procedure, and compare them to the behavior found in humans. To the best of our knowledge, we are the first to comprehensively compare inferential strategies employed by large language models and humans. We analyze the models' output both quantitatively and qualitatively via manual inspection, to provide insights into the soundness of their verbalized reasoning strategies. Our findings reveal that:

\begin{itemize}[itemsep=1pt]
    \item All models exhibit inferential strategies akin to those observed in human reasoning, such as \textit{supposition following} and \textit{chain construction}.
    \item The inferential strategy employed is significantly influenced by the model family, as different families favor different approaches.
    \item Models are often right but for the wrong reasons: the \textit{accuracy} of a model, that is the number of correct final conclusions, does not reflect whether its reasoning is \textit{sound}, i.e.\ logically follows from the statements at hand.
    \item The strategy employed by a model is closely related to the \textit{soundness} of its reasoning, where certain strategies lead to correct reasoning and others tend to introduce errors.
    \item In contrast to humans, models occasionally adopt a \textit{symbolic strategy}, where formal logical calculus is employed to solve the propositional logic problem at hand.
\end{itemize}
 
Through this work, we hope to advance the understanding of reasoning in LLMs.

\section{Strategies in Propositional Reasoning}\label{sec:Strategies-Propositional-Reasoning}
Propositional logic studies the relationships among statements (or propositions) and the methods for constructing logical arguments based on them \cite{hurley_concise_2011}. At the core of propositional logic are simple statements that can be combined through the use of logical connectives such as "not", "and", "or", and "if... then...", thereby forming more complex compound statements. Conclusions are logically deduced, where the truth value of the propositions necessitates the truth of the conclusion. This form of logical reasoning allows us to construct sound arguments that are invariably true, given the truth value of the information provided. As such, propositional logic is fundamental to various disciplines, including science, mathematics, and philosophy, where it offers a structured approach to reasoning and argumentation.

To gain insights into the inferential processes humans employ in propositional reasoning, \citet{van_der_henst_strategies_2002} conducted a series of experiments that study the behavior of participants during propositional reasoning. They formulated straightforward propositional logic problems with neutral content (the presence or absence of colored marbles in a box, similar to the problem illustrated in Figure \ref{fig:model_strategies_example}) and requested participants to articulate their thought processes while engaging with these problems. Participants were permitted the use of paper and pencil for their workings. Both their verbal explanations and written responses were meticulously recorded, transcribed and analyzed thereafter.  \citet{van_der_henst_strategies_2002} discovered five strategies reasoners commonly utilize to navigate the problems, offering insights into their inferential mechanisms employed during propositional reasoning. In the following, we give a short description of each strategy (illustrated in Figure \ref{fig:inf_strategies_human_examples}). For more details and additional examples, we refer to the original study by \citet{van_der_henst_strategies_2002}.

\begin{figure*}[htbp]
  \centering
  \begin{tikzpicture}[
    font=\fontsize{8}{10}\selectfont, % Sets the font size to 8pt
    box/.style={draw, rectangle, rounded corners, align=left, anchor=north west, minimum height=6.53cm, inner ysep=6pt, text width=0.175\textwidth},
    red box/.style={box, fill=red!20},
    green box/.style={box, fill=green!20, minimum height=3.18cm, text width=0.27\textwidth},
    yellow box/.style={box, fill=yellow!20, minimum height=3.18cm, text width=0.27\textwidth},
    orange box/.style={box, fill=orange!20, minimum height=3.18cm,  text width=0.27\textwidth},
    blue box/.style={box, fill=mylightblue, minimum height=3.18cm,  text width=0.27\textwidth},
    violet box/.style={box, fill=blue!20},
    gray box/.style={draw, rectangle, fill=gray!5, rounded corners, minimum height=1cm, text width=0.98\textwidth}
  ]

  % Gray Box
  \node[gray box] (box0){
  \textbf{Problem:}\\Evaluate whether the given conclusion logically follows from the premises below:

    % Customizing the enumerate list
    \begingroup
    \setlist[enumerate]{itemsep=0pt, parsep=0pt, partopsep=0pt, topsep=0pt, leftmargin=*} % Removes space between list items and adjusts list margins
    \begin{enumerate}
        \item There is a blue marble if and only if there is a white marble.
        \item Either there is a white marble or else there is a red marble, but not both.
        \item There is a red marble if and only if there is a pink marble.
    \end{enumerate}
    \endgroup

    Conclusion: If there is a blue marble then there is a pink marble.
  };

  % Red Box
  \node[red box, below=0.01\textwidth of box0.south west, anchor=north west] (box1) {
    \textbf{Incremental Diagram:}\\
    Blue iff white:\\[1mm] % Increase space before the diagram
    \begin{tabular}{@{}c@{}}
        White \\
        $\updownarrow$ \\
        Blue
    \end{tabular}\\[3mm] % Increase space after the diagram
    White xor red:\\[1mm]
    \begin{tabular}{@{}c|c@{}}
        White & Red \\
        $\updownarrow$ & \\ % Use \textcolor{white}{.} to add an invisible character if needed for alignment
        Blue &
    \end{tabular}\\[3mm]
    Red iff pink:\\[1mm]
    \begin{tabular}{@{}c|c@{}}
        White & Red $\leftrightarrow$ Pink \\
        $\updownarrow$ & \\
        Blue &
    \end{tabular}\\[3mm]

    If blue then not pink.\\
    Conclusion: False
  };

  % Green Box
  \node[green box, right=0.01\textwidth of box1.north east, anchor=north west] (box2) {
    \textbf{Supposition Following:}\\
    \textit{Assuming} we have a blue marble.\\
    Then there is a white marble.\\
    This means there is no red marble.\\
    Thus there can not be a pink marble.\\
    \vspace{2mm}
    If blue then not pink.\\
    Conclusion: False
    \vspace{2.5mm}
  };

  % Yellow Box
  \node[yellow box, right=0.01\textwidth of box1, below=0.01\textwidth of box2.south west, anchor=north west] (box3) {
    \textbf{Chain Construction:}\\
    If blue then white.\\
    If white then not red.\\
    If not red then not pink.\\
    \vspace{2mm}
    Therefore, if blue then not pink.\\
    Conclusion: False
    \vspace{6mm}
  };

  % Orange Box
  \node[orange box, right=0.01\textwidth of box2.north east, anchor=north west] (box4) {
    \textbf{Compound Strategy:}\\
    1. Blue iff white.\\
    2. White xor red.\\
    \vspace{-1.5mm}
    \begin{tikzpicture}
    \draw[dashed] (0,0) -- (1.2, 0);
    \end{tikzpicture}\\
    If blue then not red.\\
    3. Red iff pink.\\
    \vspace{-1.5mm}
    \begin{tikzpicture}
    \draw[dashed] (0,0) -- (1.2,0);
    \end{tikzpicture}\\
    If blue then not red and not pink.\\
    Conclusion: False
  };

  % Blue Box
  \node[blue box, right=0.01\textwidth of box3, below=0.01\textwidth of box4.south west, anchor=north west] (box5) {
    \textbf{Concatenation Strategy:}\\
    Blue iff white.\\
    White xor red.\\
    Red iff pink.\\
    \vspace{2mm}
    Blue iff (white xor (red iff pink)).\\
    \vspace{2mm}
    Therefore, blue xor pink.\\
    Conclusion: False
    \vspace{0.5mm}
  };

  \draw [thick, dashed] ([xshift=0.01\textwidth]box4.north east) -- ([xshift=0.01\textwidth]box5.south east);

  % Violet Box
  \node[violet box, right=0.02\textwidth of box4.north east, anchor=north west] (box6) {
    \textbf{Symbolic Strategy:}\\
    Blue iff white.\\
    White xor red.\\
    Red iff pink.\\[1\baselineskip]

    First, let's write down the statements in a logical form:
    \begingroup
    \setlist[enumerate]{itemsep=0pt, parsep=0pt, partopsep=0pt, topsep=0pt, leftmargin=*} % Removes space between list items and adjusts list margins
    \begin{enumerate}
        \item $B \leftrightarrow W$
        \item $W\oplus R$
        \item $R \leftrightarrow P$
        \item[]
    \end{enumerate}
    \endgroup
    Now, let's derive the conclusion using these statements [...].
    \vspace{9.5mm}
  };

\end{tikzpicture}
  \caption{An example for each of the five inferential strategies identified by~\citet{van_der_henst_strategies_2002} (to the left of the dashed vertical line) that human reasoners employ when solving tasks of propositional logic. Each strategy is illustrated by a single example adopted from the transcribed recordings published by the original study. In addition, we provide an example of the \symbolicstrat~\textit{symbolic strategy} occasionally encountered in LLMs (to the right of the dashed line). "Iff" denotes a biconditional, while "xor" indicates an exclusive disjunction.}
  \label{fig:inf_strategies_human_examples}
\end{figure*}

\vspace{2mm}

\noindent\textbf{Incremental Diagram.} This strategy involves the creation of a comprehensive diagram that keeps track of all potential outcomes compatible with the premises of the problem. During the reasoning process, individuals progressively \textit{increment} their diagrams to incorporate new information derived (see left box in Figure \ref{fig:inf_strategies_human_examples}). The result is a single diagram that records a variety of possibilities compatible with the premises, often including even those that might be irrelevant to the task.\footnote{In contrast to~\citet{van_der_henst_strategies_2002}, we in fact observe no single occurrence of the \textit{incremental diagram strategy} in LLMs, despite the authors finding that this strategy is most frequently employed by humans. We believe that this discrepancy stems from the use of pen and paper in human assessments, implicitly encouraging diagrammatic reasoning. Exploring how this observation changes with vision-language models would be an intriguing area for future research.}

\vspace{1mm}

\noindent\textbf{Supposition Following.} Reasoners employing this strategy start with a supposition, e.g. by assuming a marble of a certain color. Subsequently, they trace the implications of that supposition, logically following from the premises at hand, as illustrated in the upper second box from the left of Figure \ref{fig:inf_strategies_human_examples}. The result is a sequence of literals (in this case, marbles of a certain color) without logical connectives. The efficiency and success of \textit{supposition following} strongly depends on the supposition made by the reasoner. While some suppositions lead to inferences that are relevant to the problem, others might lead to irrelevant conclusions.

\vspace{1mm}

\noindent\textbf{Chain Construction.} When employing this strategy, reasoners construct a chain of conditional statements derived either from the premises in the problem description or from intermediate deductions. An example of \textit{chain construction} is displayed in the lower second box from the left of Figure \ref{fig:inf_strategies_human_examples}. Premises are converted into a chain of conditional statements that are linked by their entities. A distinctive feature of this \textit{chain} is the interconnection between conditionals, where the consequent of one conditional is the antecedent of the following.

\vspace{1mm}

\noindent\textbf{Compound Strategy.} Reasoners following the \textit{compound strategy} combine two or more statements to derive a new \textit{compound conclusion}. This process yields a series of novel conclusions, each building upon the preceding ones. An illustrative example of this strategy is given in the upper second box from the right of Figure \ref{fig:inf_strategies_human_examples}. Based on the first two premises, the compound conclusion: \textit{``If blue then not red.''} is inferred, and then used to draw another compound conclusion (\textit{``If blue then not red and not pink.''}) together with the last premise of the problem statement.
 
\vspace{1mm}

\noindent\textbf{Concatenation Strategy.} This approach entails the concatenation of two or more statements into a \textit{single conclusion} encompassing the logical implications of each combined proposition. This strategy is subtle and has only been infrequently observed by \citet{van_der_henst_strategies_2002}. An example of the strategy is illustrated in the lower second box from the right of Figure \ref{fig:inf_strategies_human_examples}.

\vspace{1mm}

\noindent\textbf{Symbolic Strategy.} We could identify an additional strategy occasionally employed by LLMs, which has not been observed by \citet{van_der_henst_strategies_2002} in human reasoners. This strategy, which we denote as \textit{symbolic strategy}, is characterized by models employing formal logical calculus to solve the tasks at hand. When following this strategy, models either translate logical statements that are expressed in natural language (e.g. \textit{``If there is a white marble then there is not a red marble.''}) into formal logic ($W \rightarrow \neg R$), and then operate on those expressions, or create a truth table from which they aim to infer the validity of the conclusion. An illustration of this strategy is provided in the right box of Figure \ref{fig:inf_strategies_human_examples}.

\section{Experimental Setup}\label{sec:Method}

\begin{figure*}[t!]
  \centering
  \input{tikz_files/appendix/C-Qualitative-Examples/Chain-Construction/problem1_llama2_70B_chain_construction}
  \caption{The response (lower left box) of \LLaMABig{} to problem \hyperref[fig:appendix_a_task_prompt]{1} (top box) of the problem set, demonstrating \chainstrat~\textbf{chain construction}. The model correctly constructs a chain of conditionals (highlighted in yellow within the model's response) based on the premises, leading from the antecedent of the final conclusion to its consequent. Comments made by the annotators are presented in the adjacent right panel.}
  \label{fig:chain_construction_example_llama2_70B_problem1}
\end{figure*}

\noindent\textbf{Task Overview.} Our task setup aligns with the experiment conducted by \citet{van_der_henst_strategies_2002} to allow for a fair comparison between the inferential strategies found in humans and those identified in LLMs.\footnote{More specifically, experiment one of \citet{van_der_henst_strategies_2002}.} In particular, we evaluate each model on the 12 problems of propositional logic suggested by \citet{van_der_henst_strategies_2002} (an overview of each problem can be found in Figure \ref{fig:appendix_a_task_prompt} in the appendix). For each problem, models are presented with a set of statements (or premises) and must determine whether a given conclusion logically follows (for an example, see Figure \ref{fig:model_strategies_example}). Eight out of 12 problems involve three premises and a conclusion, while the remaining four problems consist of four premises leading to a conclusion. All premises, as well as the conclusions resemble either biconditionals, exclusive disjunctions or conditionals. Two problems (4 and 6) include a redundant first premise. All premises are stated such that two subsequent statements contain one proposition in common, except of two problems (11 and 12), which are arranged in a non-sequential manner. For half of the problems, the conclusions logically follow from the premises, whereas for the other half, they do not. To avoid the influence of external knowledge and ensure content neutrality, \citet{van_der_henst_strategies_2002} framed the problems around the presence of colored marbles in a box, with colors assigned randomly to each entity within a problem.

\vspace{1mm}

\noindent\textbf{Language Models.} We aim to investigate various factors that might impact the inferential strategies displayed by LLMs. These factors include the type of model, its size, and the emphasis on alignment during training \cite{tunstall_zephyr_2023}. Therefore, we assess a total of five models, consisting of three prominent open-access model types: Llama 2 \cite{touvron_llama_2023} with model sizes of 7B, 13B, and 70B, the recently released Mistral-7B model \cite{jiang_mistral_2023}, and Zephyr-7B \cite{tunstall_zephyr_2023}, an extension of Mistral-7B with a focus on intent alignment through fine-tuning with AI Feedback (AIF). For our evaluations, we utilize the publicly accessible model weights from the HuggingFace platform, specifically \texttt{Llama-2-chat-hf}\footnote{\label{fn:llama2}\href{https://huggingface.co/meta-llama}{https://huggingface.co/meta-llama}}(7B, 13B, and 70B), \texttt{Mistral-7B-Instruct-v0.2},\footnote{\href{https://huggingface.co/mistralai/Mistral-7B-Instruct-v0.2}{https://huggingface.co/mistralai/Mistral-7B-Instruct}} and \texttt{zephyr-7b-beta}.\footnote{\href{https://huggingface.co/HuggingFaceH4/zephyr-7b-beta}{https://huggingface.co/HuggingFaceH4/zephyr-7b-beta}} We consciously opt not to include proprietary models accessible via paid APIs, despite their reported superior performance in reasoning tasks \cite{gemini_team_gemini_2023}. This methodological choice reflects our commitment to promoting transparent and reproducible scientific research. Note that in this work, we refer to the above models when using abbreviations such as LLaMA-2, Mistral-7B-Instruct or Zephyr-7B-$\beta$.

\vspace{1mm}

\noindent\textbf{Evaluation Setup.} We prompt each model with a system message providing context about the task they are about to solve and the format in which they should answer (for the full prompt, see Figure \ref{fig:appendix_a_task_prompt} in the appendix). Analogous to \citet{van_der_henst_strategies_2002}, we inform the model of its participation in an experiment designed to explore reasoning processes, and instruct it to \textit{``think aloud''} as it tackles the problem. In addition to the system message, we provide a user prompt that contains the problem description. In cases where the model does not accept system messages (such as \texttt{Mistral-7B-Instruct-v0.2}), we prepend the content of the system message to the user prompt. To prevent biasing the model towards a certain strategy, we refrain from providing few-shot examples, as done also by~\citet{leidinger-etal-2023-language}. Instead, we elicit reasoning through zero-shot chain-of-thought prompting (\textit{``Let's think step by step''}) \cite{kojima_large_2022}. Answers for each model are generated with nucleus sampling using \texttt{Llama-2-chat-hf}'s default values ($\text{top-}p = 0.9$, temperature $T = 0.6$), as we found this configuration to work well for all models. To account for the statistical nature of language models, we ask each model to solve the set of propositional problems across 5 random seeds, resulting in a total of 60 responses per model. Our code is publicly available at: \href{https://github.com/mainlp/inferential-strategies}{https://github.com/mainlp/inferential-strategies}.

We record all answers and manually evaluate them (a total of 300 responses) for strategies employed in their reasoning (see Figure \ref{fig:chain_construction_example_llama2_70B_problem1} for an example). For each model response, we qualitatively evaluate for \textit{strategy} and \textit{soundness}. That is, we manually label the inferential strategies identified, and the logical validity of the model's reasoning. In addition, we record whether the final answer is correct. In cases of faulty reasoning, we categorize the type of error. This comprehensive manual evaluation of model responses is independently conducted by two hired students with expertise in manual data annotation. To gauge the quality of the annotations, we report an overall Cohen's Kappa value of $\kappa = 0.98$. For details on the inter-annotator agreement of each label, we refer to Table \ref{tab:cohens_kappa} in the appendix. Further annotated examples can be found in Appendix \ref{sec:appendix_c_annotated_model_responses}. Following the recommendations put forward by~\citet{leidinger-etal-2023-language}, we make all input prompts, model responses and manual annotations publicly available at: \href{https://huggingface.co/datasets/mainlp/inferential\_strategies}{huggingface.co/datasets/mainlp/inferential\_strategies}.

% Relative occurences of strategies found in the models' reasoning
\begin{table*}[htp]
\centering
\renewcommand{\arraystretch}{1.2}
{\fontsize{8}{10}\selectfont
\begin{tabular}
{>{\centering\arraybackslash}m{2.4cm} *{5}{>{\centering\arraybackslash}m{1.4cm}} !{\vrule width 1pt} *{2}{>{\centering\arraybackslash}m{1.4cm}}}
\toprule
Model & \cellcolor{green!20} Supposition Following & \cellcolor{yellow!20} Chain Construction & \cellcolor{orange!20} Compound Conclusion & \cellcolor{mylightblue} Concatenation Strategy & \cellcolor{blue!20} Symbolic Strategy & Correct Answer & Sound Reasoning \\ \hline
\midrule
Zephyr-7B-$\beta$ & $\mathbf{60.0\%} \, (\textit{55.1})$ & $18.3 \% \, (\textit{17.3})$ & $10.0 \% \, (\textit{8.9})$ & $1.7 \% \, (\textit{1.4})$ & $20.0 \% \, (\textit{17.3})$ & $45.0 \pm 15.5$ & $25.0 \pm 10.5$ \\
\cmidrule{1-8}
Mistral-7B-Instruct & $\mathbf{35.0} \% \, (\textit{38.4})$ & $10.0 \% \, (\textit{10.7})$ & $\mathbf{35.0} \% \, (\textit{38.4})$ & $3.3 \% \, (\textit{3.4})$ & $8.3 \% \, (\textit{9.1})$ & $55.0 \pm 10.0$ & $25.0 \pm 7.5$ \\
\cmidrule{1-8}
\LLaMASmall{} & $\mathbf{20.0 \%} \, (\textit{50.2})$ & $\mathbf{20.0 \%} \, (\textit{30.2})$ & $6.7 \% \, (\textit{10.9})$ & $3.3 \% \, (\textit{5.4})$ & $1.7 \% \, (\textit{3.3})$ & $46.7 \pm 6.7$ & $0.0 \pm 0.0$ \\
\LLaMAMedium{} & $28.3 \% \, (\textit{35.7})$ & $\mathbf{36.7 \%} \, (\textit{46.9})$ & $6.7 \% \, (\textit{8.7})$ & $6.7 \% \, (\textit{8.7})$ & $0.0 \% \, (\textit{0.0})$ & $40.0 \pm 8.2$ & $15.0 \pm 6.2$ \\
\LLaMABig{} & $45.0 \% \, (\textit{42.3})$ & $\mathbf{50.0 \%} \, (\textit{46.8})$ & $3.3 \% \, (\textit{2.9})$ & $1.7 \% \, (\textit{1.8})$ & $6.7 \% \, (\textit{6.2})$ & $56.7 \pm 6.2$ & $31.7 \pm 9.7$ \\
\cmidrule{1-8}
\cmidrule[\heavyrulewidth]{1-8}
Human Reasoner$^{\dagger}$ & $- \ (21.0)$ & $-  \ (25.0)$ & $-  \ (19.0)$ & $-  \ (0.0)$ & $-  \ (0.0)$ & $100 \pm 0.0$ & $-$ \\
\bottomrule
\end{tabular}
}
\caption{Relative occurrences of inferential strategies employed by the different language models when solving the problems of propositional logic. All values reflect average percentages, calculated over five random seeds, with standard deviations reported in Table \ref{tab:inf_strategies_frequencies_with_stdev} in the appendix. Strategies that a model favors are highlighted in bold. Values in parentheses denote fractions with respect to the total number of strategies employed by that model. Values of correct answers and instances of sound reasoning are reported with their standard deviations. $^{\dagger}$The comparison with human reasoners is based on findings by \citet{van_der_henst_strategies_2002}, where dashes denote missing values.}
\label{tab:inf_strategies_frequencies}
\end{table*}

\section{Results and Analysis}\label{sec:Results}
In this section, we present the results of our evaluation. We begin with a quantitative analysis of the inferential strategies employed by LLMs, as well as the logical validity of their reasoning. This is followed by a qualitative analysis providing a more in-depth examination of the models' reasoning.

\subsection{Quantitative Analysis}\label{subsec:quantitative_analysis}
Table \ref{tab:inf_strategies_frequencies} provides an overview of the frequencies with which large language models employ inferential strategies when navigating the problems of propositional logic described in Section \ref{sec:Method}. Our evaluation reveals that all models display strategies akin to those observed by \citet{van_der_henst_strategies_2002}. In particular, we find that, similar to humans, models commonly employ \textit{supposition following}, \textit{chain construction} and the \textit{compound strategy}. In addition, we observe that models occasionally utilize the \textit{symbolic strategy}, employing techniques from logical calculus to solve the tasks (see Section \ref{sec:Strategies-Propositional-Reasoning}). Note that, similar to humans, models might switch from one strategy to another during a single problem, demonstrating multiple strategies within their responses (see Figure \ref{fig:appendix_c_chain_symbolic_strategy_zephyr_7B_problem6} in the appendix for an example). Surprisingly, we observe that distinct model families favor different inferential strategies. For instance, Zephyr-7B-$\beta$ predominantly employs \textit{supposition following}, while Mistral-7B-Instruct is equally inclined towards drawing \textit{compound conclusions}. In contrast, models from the Llama 2 series tend to rely on \textit{supposition following} and \textit{chain construction}, with negligible use of the \textit{compound strategy}. Our analysis further reveals a discrepancy between the correctness of the models' final answers and the logical soundness of their reasoning. While all models achieve an answer accuracy that approximately coincides with chance in our experimental setup, an analysis of their reasoning validity reveals a different picture: \LLaMABig{} outperforms the other models by reasoning correctly in about 31.7\% of cases, while Zephyr-7B-$\beta$ and Mistral-7B-Instruct produce sound reasoning in 25\% of the problems. We note that all models perform rather poorly on the propositional tasks, with \LLaMASmall{} failing entirely to construct sound arguments.

\noindent\textbf{Human Reasoning.} \citet{van_der_henst_strategies_2002} compute the percentages with which human reasoners employ inferential strategies with respect to the total number of strategies observed in their experiment, and not with respect to the total number of problems considered. Thus, their reported values mainly reflect which strategies are favored more or less by the reasoner, but do not provide information about how frequently a strategy has been observed in the overall context. To make our findings comparable to the results of \citet{van_der_henst_strategies_2002}, we convert our results respectively (see values in parentheses in Table \ref{tab:inf_strategies_frequencies}). We note that almost all models seem to favor \textit{supposition following} to a higher degree than human reasoners, who employ this strategy in only about 21\% of overall use. In contrast, humans seem to draw compound conclusions more readily, except for Mistral-7B-Instruct which shows a tendency more than twice as high. Overall, both LLMs and humans hardly employ the \textit{concatenation strategy}. Interestingly, \citet{van_der_henst_strategies_2002} report that all reasoners successfully solve the problems of propositional logic, though not always for the correct reasons. While the study does not provide data on the number of problems where humans reasoned correctly, the high success rate of human participants contrasts sharply with the performance of the models.

 \begin{figure*}[tbp]
  \centering
  \includegraphics[width=0.85\textwidth]{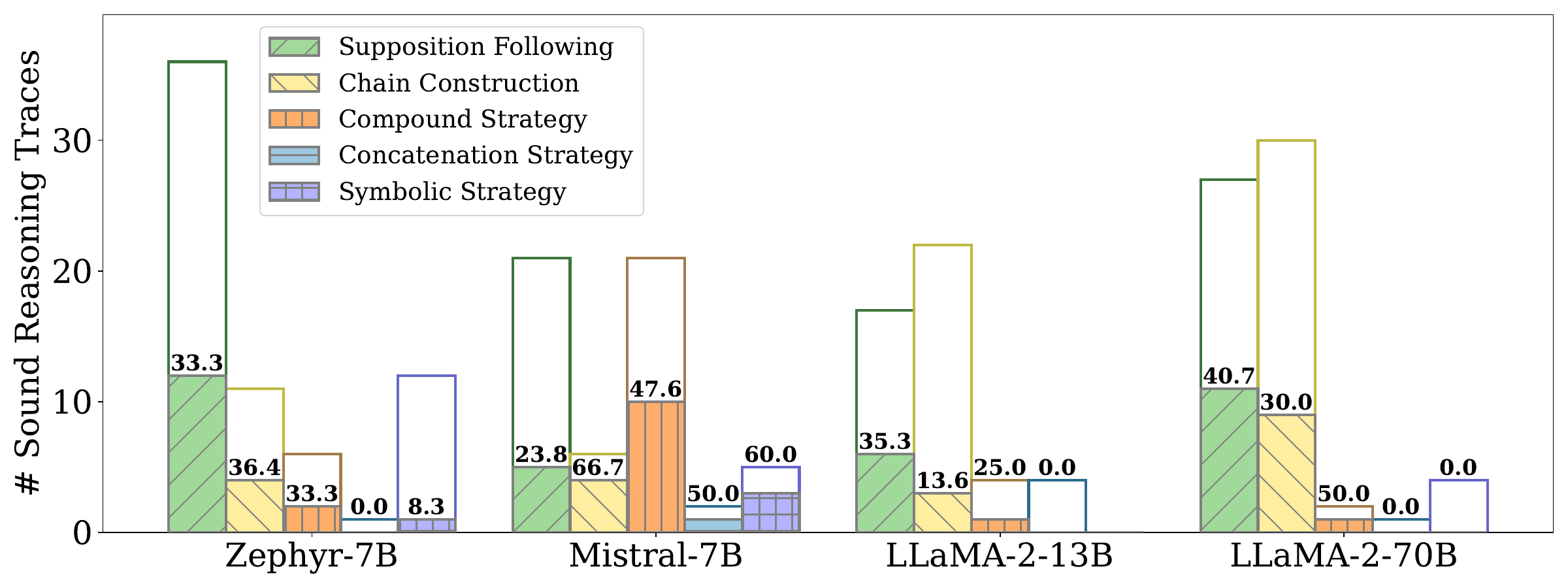}
  \caption{Instances where models generate sound reasoning traces that logically follow from the problem statement. For each inferential strategy, the ratio of sound reasoning traces (represented by the filled portion) to the overall application of that strategy (denoted by the unfilled bar) is depicted. Ratios are expressed as percentages above the corresponding filled section. Note that \LLaMASmall{} is not displayed as it does not exhibit sound reasoning.}
  \label{fig:reasoning_soundness_strategies}
\end{figure*}

\vspace{1mm}

\noindent\textbf{Effect of Model Size.} Our evaluation of the Llama 2 series across three different model sizes—7B, 13B, and 70B parameters—demonstrates that model scale significantly influences the frequency with which strategies are employed by the model. In particular, we observe that with increasing model size, Llama 2 employs strategies more readily. Furthermore, larger models within the Llama 2 framework are observed to generate a greater number of sound reasoning traces. We interpret this trend as a result of the model's improving proficiency in strategic reasoning as its scale increases.

\vspace{1mm}

\noindent\textbf{Effect of Alignment.} The alignment of a model's response with human preferences is crucial to emulate human-like behavior \cite{ouyang_training_2022}. Zephyr-7B-$\beta$ is an iteration of Mistral-7B that is fine-tuned with AI Feedback (AIF) for improved intent alignment \cite{tunstall_zephyr_2023}. In comparison to the observations made by \citet{van_der_henst_strategies_2002}, where besides the \textit{incremental diagram strategy} (34\%), \textit{chain construction} was employed most frequently by humans,  Zephyr-7B-$\beta$ demonstrates a marked preference for \textit{supposition following} and significantly less engagement in \textit{chain construction}. Moreover, it is noteworthy that among the evaluated models, Zephyr-7B-$\beta$ most frequently adopts the \textit{symbolic strategy}, an approach not reported in human reasoners. 

\vspace{1mm}

\noindent\textbf{Sound Reasoning.} As previously highlighted, the accuracy of a model's final answer does not necessarily serve as a reliable indicator of its reasoning capability. In particular, we observe that models often arrive at correct answers, but through flawed reasoning processes (refer to Figure \ref{fig:appendix_c_chain_construction_llama2_70B_problem7} in the appendix for an illustration). Interestingly, we also find instances where models provide incorrect final answers despite reasoning correctly (for an example, see Figure \ref{fig:appendix_c_compound_supp_strategy_mistral_instruct_problem8} in the appendix). Our analysis reveals only a moderate positive correlation between the accuracy of the models' final answers and the logical soundness of their reasoning, with a Pearson correlation coefficient $r(298) = 0.45$ and a statistically significant p-value of less than $0.0001 (p = 1.6 \times 10^{-16})$. This observation aligns with findings from previous studies \cite{NEURIPS2022_c4025018, creswell2022faithful} and underscores the need for more nuanced evaluation procedures, particularly in multiple-choice settings, where models might select the correct answer by chance rather than through rigorous reasoning.

In Figure \ref{fig:reasoning_soundness_strategies}, we explore the relationship between the inferential strategies employed by the models and the validity of their reasoning. For each strategy, we quantify the proportion of instances where the models' reasoning is sound, compared to the overall application of that strategy. Our analysis reveals variability in the effectiveness with which different models apply various strategies. For example, Mistral-7B-Instruct tends to reason correctly when using approaches such as the \textit{chain}, \textit{compound}, or \textit{symbolic strategy}, yet frequently encounters reasoning errors with \textit{supposition following}. On the other hand, \LLaMABig{} exhibits proficiency in \textit{supposition following}, but struggles with the \textit{symbolic strategy}.

\subsection{Qualitative Analysis}\label{subsec:qualitative_analysis}
We supplement our quantitative analysis by a more detailed qualitative analysis of the models' reasoning behavior. Figure \ref{fig:chain_construction_example_llama2_70B_problem1} depicts \LLaMABig's response to problem \hyperref[fig:appendix_a_task_prompt]{1} of the task set. The response illustrates a frequently observed behavior. Initially, models tend to analyze the problem's propositions, often by paraphrasing each premise and the conclusion to be evaluated. They then embark on a reasoning process, typically utilizing one of the previously mentioned strategies. In the example, \LLaMABig{} employs \textit{chain construction}, creating a logical chain of conditionals that leads from the antecedent of the final conclusion to its consequent, thereby correctly affirming the conclusion's logical validity. A notable pitfall in such reasoning chains is the models' occasional misinterpretation of logical negations, leading to erroneous chains like: A $\rightarrow \neg$ B; B $\rightarrow $ C; therefore A $\rightarrow$ C, where the negation in the first conditional is overlooked (for an illustrative case, refer to Figure \ref{fig:appendix_c_chain_construction_llama2_13B_problem10} in the appendix). This behavior can be found across all models and aligns with previous work reporting difficulties of LLMs in understanding logical negations \cite{truong-etal-2023-language}.

When employing \textit{supposition following}, models often fail to consider all implications of their assumptions. Instead, they tend to focus only on immediate inferences, while overlooking further consequences crucial for assessing the conclusion's validity.  This leads to models prematurely concluding the inability to definitively determine the logical validity of the final conclusion: \textit{``Based on our analysis, we cannot definitively say that the conclusion logically follows from the given statements''} (see Figure \ref{fig:appendix_c_supposition_following_mistral_instruct_problem7} in the appendix for a respective example). Another source of error in \textit{supposition following} involves models making improper suppositions, such as conjecturing about a marble not mentioned in the final conclusion, and deriving disjointed intermediate conclusions that do not aid in solving the problem. An example of this behavior can be found in Figure \ref{fig:appendix_c_supposition_following_mistral_instruct_problem9} in the appendix.

Finally, we identify two behaviors in models that mirror logical errors seen in human reasoners \cite{van_der_henst_strategies_2002}. First, models frequently attempt to prove an exclusive disjunction (A $\oplus$ B) by only considering a single conditional case (A $\rightarrow \neg$ B), and second, they sometimes engage in the logical fallacy known as denial of the antecedent: A $\rightarrow$ B; therefore $\neg$ A $\rightarrow \neg$ B (for illustrative examples, see Figures \ref{fig:appendix_c_chain_construction_llama2_70B_problem5} and \ref{fig:appendix_c_chain_construction_llama2_70B_problem12} in the appendix, respectively).

\section{Related Work}\label{sec:Related-Work}
\noindent\textbf{Human Strategies in Deductive Reasoning.} A considerable amount of research, especially within psychology and cognitive science, has explored how humans approach deductive reasoning tasks \cite{schaeken_deductive_2000}. A prominent focus of these studies is on heuristics, which are cognitive shortcuts that individuals employ to arrive at satisfactory conclusions in deductive reasoning despite potential flaws in the underlying logic \cite{kahneman_judgment_1982, evans_bias_1989, gigerenzer_simple_1999, davis_biases_2018}. For instance, \citet{woodworth_atmosphere_1935} demonstrate that individuals tend to accept conclusions in syllogistic reasoning as valid when they share logical quantifiers with the premises, regardless of their actual logical validity. Nonetheless, such reliance on heuristics can result in errors and falls short of the level of strategic reasoning necessary to develop sound and coherent arguments \cite{kahneman_thinking_2012}. Further research has delved into more sophisticated strategies utilized by individuals in deductive reasoning. Based on the mental model theory \cite{johnson-laird_mental_1986}, \citet{bucciarelli_strategies_1999} identify a variety of strategies commonly employed by individuals in syllogistic reasoning. \citet{byrne_reasoning_1997} study strategies of individuals in \textit{knight-and-knave} puzzles, where the truthfulness of statements made by hypothetical characters have to derived. Their experiments reveal that humans engage in both forward and backward inferences to navigate through potential solutions.

\vspace{1mm} 

\noindent\textbf{Human Reasoning Behavior in LLMs.} Recent research has started to explore the extent to which LLMs mirror human-like reasoning behaviors. \citet{dasgupta_language_2023} demonstrate content-effects akin to those observed in human reasoning, where the deductive process is influenced by the content of the problem statement. \citet{eisape_systematic_2023} find that LLMs, similar to humans, exhibit biases such as ordering effects in syllogistic reasoning tasks. Several other studies have delved into the prevalence of biases and heuristics within LLMs \cite{binz_using_2023, talboy_challenging_2023, shaki_cognitive_2023, suri2024large}. However, to the best of our knowledge, we are the first who study the presence of more sophisticated human strategies in the context of LLM-based deductive reasoning.

\vspace{1mm}

\noindent\textbf{Faithful Reasoning with LLMs.} Large language models can be instructed to explain the reasoning process by which they derive their final conclusions \cite{NEURIPS2022_9d560961, kojima_large_2022}. However, several studies indicate that these self-explanations might not always be \emph{faithful}, i.e.\ accurately represent the model's underlying reasoning process \cite{jacovi-goldberg-2020-towards, agarwal2024faithfulness, 10.1162/coli_a_00511}. For instance, \citet{NEURIPS2023_ed3fea90} demonstrate that LLMs such as GPT-3.5 \cite{GPT3point5} and Claude 1 \cite{anthropic2023claude1} often fail to mention biasing features in their input that significantly influence their decisions. Instead, the models produce \emph{plausible} yet misleading explanations that give a false account of the underlying decision process. \citet{lanham2023measuring} probe the faithfulness of explanations by evaluating how the final conclusions of LLMs change when rationales are truncated or errors are introduced. Their findings reveal that the extent to which models rely on their rationales varies strongly across models and tasks. \citet{matton2024walk} propose a method to quantify the faithfulness of explanations based on high-level concepts in the models' input that influence decision-making. By measuring the difference between the set of concepts that LLMs deem influential and the set that truly are, instances of unfaithfulness could be identified, including cases where LLMs overlook the impact of social biases in their decision-making processes. Another important consideration is whether the model's final conclusion aligns with its preceding explanation. As highlighted in Section \ref{subsec:quantitative_analysis}, a correct conclusion might not always be the product of a logically sound reasoning trace, particularly in multiple-choice setups. Conversely, a sound rationale may not always lead to a logically consistent answer. Related work by \citet{NEURIPS2022_c4025018} indicates that in question-answering and natural language inference tasks, explanations generated by LLMs such as OPT \cite{zhang2022opt} and GPT-3 \cite{NEURIPS2020_1457c0d6} often do not entail the models' final conclusions. Further studies aim to enhance the models' faithfulness, for instance by enforcing causality from proof generation to entailment prediction. This can be achieved by either restricting the model's context \cite{sanyal-etal-2022-fairr, creswell2022faithful, radhakrishnan2023question}, or by utilizing deterministic tools that are inherently faithful by design \cite{osti_10463284}.

\section{Conclusion}\label{sec:Conclusion}
In this paper, we examine the inferential strategies employed by LLMs in solving problems of propositional logic. Through a comprehensive evaluation of their reasoning behavior, we demonstrate that LLMs adopt strategies akin to those observed in human reasoners. Our quantitative analysis reveals that the frequency with which a model adopts a specific strategy strongly depends on its type, size, and fine-tuning procedure. Moreover, our analysis suggests that the accuracy of a model's final conclusions does not adequately capture its reasoning capabilities, underscoring the importance of a more sophisticated evaluation framework that includes the model's reasoning paths. We also provide a qualitative analysis of typical reasoning behaviors among models, pinpointing prevalent errors such as difficulties in understanding negations or recognizing all implications of a supposition.

\section{Limitations}\label{sec:Limitations}
While our work contributes to the understanding of reasoning processes in large language models by demonstrating that these models employ inferential strategies in propositional logic similar to humans, it encompasses several limitations that could be addressed in future work.

\vspace{1mm}

\noindent\textbf{Task setup.} Our study is constrained by a limited set of problems, designed within a fixed framework that revolves around hypothesis validation based on 3-4 statements of propositional logic. We employ a constant and neutral content, disregarding potential content-effects on the models' reasoning behavior, as shown by \citet{dasgupta_language_2023}. Similarly, we have not yet examined factors such as the complexity of the problems, the differences between hypothesis validation and generation, and the impact of logical connectives utilized in the premises. We believe that these factors are worth investigating and leave a detailed examination to future work.

\vspace{1mm}

\noindent\textbf{Evaluation Framework.} The extent of our manual evaluation is limited by both the number of samples reviewed and the quantity of annotators involved. Despite our efforts to maximize the use of available resources, these constraints may affect the scalability and reliability of our results. Additionally, we instruct all models through zero-shot chain-of-thought prompting (\textit{``Let's think step by step''}) \cite{kojima_large_2022}. Exploring alternative reasoning frameworks, such as Tree of Thoughts \cite{yao_tree_2023} or Graph of Thoughts \cite{besta_graph_2024}, could provide valuable insights into their influence on model behavior and the inferential strategies adopted. Based on our annotated data, we endeavored to develop a classifier capable of automatically identifying the inferential strategies employed in the models' output, which was intended to complement our manual evaluation setup. However, due to the complexity of the task and limited size of our annotated dataset, our classifier struggled with generalization to new, unseen responses. In future endeavors, we aim to allocate more resources towards expanding our manual annotation efforts and explore this direction further. Finally, our study predominantly offers a behavioral analysis and does not delve into the mechanistic aspects that might explain the diversity in strategy usage by the models. Investigating how model-internal mechanisms might influence their choice of reasoning strategy presents a compelling direction for future research.

\section*{Acknowledgements}
We would like to thank the MaiNLP lab members for their insightful feedback, specifically Diego Frassinelli, Rob van der Goot, Siyao Peng, Robert Litschko, Jian Lan, Daniela Teodorescu, Xinpeng Wang, Verena Blaschke, Elena Senger, Elisa Bassignana, and Max Müller-Eberstein. Furthermore, we would like to express our gratitude to Huangyan Shan for her valuable work and support in data annotation. Our appreciation extends to the anonymous reviewers for their comments and suggestions. Lastly, we acknowledge the support for BP through the ERC Consolidator Grant DIALECT 101043235.

\bibliography{anthology, custom}

% ############
% # Appendix #
% ############
\appendix

\section{Additional Experimental Details}\label{sec:appendix_a_add_experimental_details}
In this section, we provide additional details about the experimental setup, including supplementary information about the problem formulations and prompts utilized.

\begin{figure*}[htbp]
  \centering
  \begin{tikzpicture}
        \definecolor{beige}{RGB}{245, 245, 220} % RGB for beige {249,228,190}
            
        % Define styles
        font=\fontsize{8}{9.6}\selectfont, % Sets the font size globally for the tikzpicture
        \tikzset{topbox/.style={rectangle, rounded corners, draw=black, fill=beige, text width=1.\textwidth, text height=0.5cm, align=left, anchor=north,}}
        \tikzset{box/.style={rectangle, rounded corners, draw=black, fill=gray!10, text width=0.3183\textwidth, text height=0.5cm, align=left, minimum height=3.5cm,}}
                
        % Top gray box
        \node[topbox] at (0,0) (topbox){
            [INST] <<SYS>>\\
            You participate in an experiment that tries to understand how people reason.\\
            Your task is to solve logical reasoning problems. In particular, you are given set of statements and your task is to say whether a conclusion logically follows from the statements.\\
            Please answer with `True' or `False' for each conclusion. In addition, it is important that you think-aloud as you tackle the problem and report every step in your reasoning process.\\
            <</SYS>> \\
            \vspace{2mm}
            \#\#\# Instruction \#\#\#\\
            Explain whether the conclusion logically follows from the set of statements below. Please report all your reasoning steps.\\
            End your reasoning with: Conclusion: True/False. \\
            \vspace{2mm}
            Statements:\\
            <statements and conclusion from below>\\
            \vspace{2mm}
            Let's think step by step. [/INST]
            \vspace{2mm}
        };

        % First row of beige boxes
        \node[box, below=0.02\textwidth of topbox.south west, anchor=north west] (box1) {
            \textbf{Problem 1:}\\
            \vspace{1mm}
            \begingroup
            \setlist[enumerate]{itemsep=0pt, parsep=0pt, partopsep=0pt, topsep=0pt, leftmargin=*} % Removes space between list items and adjusts list margins
            Statements:\\
            \begin{enumerate}
                \item White xor black.
                \item Black xor pink.
                \item Pink iff gray.
                \item[] 
            \end{enumerate}
            \endgroup
            Conclusion: If white then gray.
            \vspace{6mm}
        };
        \node[box, right=0.01\textwidth of box1] (box2) {
            \textbf{Problem 2:}\\
            \vspace{1mm}
            \begingroup
            \setlist[enumerate]{itemsep=0pt, parsep=0pt, partopsep=0pt, topsep=0pt, leftmargin=*} % Removes space between list items and adjusts list margins
            Statements:\\
            \begin{enumerate}
                \item Brown iff orange.
                \item Orange xor yellow.
                \item Yellow iff green.
                \item[] 
            \end{enumerate}
            \endgroup
            Conclusion: If brown then green.
            \vspace{6mm}
        };
        \node[box, right=0.01\textwidth of box2] (box3) {
            \textbf{Problem 3:}\\
            \vspace{1mm}
            \begingroup
            \setlist[enumerate]{itemsep=0pt, parsep=0pt, partopsep=0pt, topsep=0pt, leftmargin=*} % Removes space between list items and adjusts list margins
            Statements:\\
            \begin{enumerate}
                \item Green iff purple.
                \item If purple then gray.
                \item Gray xor yellow.
                \item[] 
            \end{enumerate}
            \endgroup
            Conclusion: Green xor yellow.
            \vspace{6mm}
        };

        % Second row of beige boxes
        \node[box, below=0.01\textwidth of box1] (box4) {
            \textbf{Problem 4:}\\
            \vspace{1mm}
            \begingroup
            \setlist[enumerate]{itemsep=0pt, parsep=0pt, partopsep=0pt, topsep=0pt, leftmargin=*} % Removes space between list items and adjusts list margins
            Statements:\\
            \begin{enumerate}
                \item Red xor maroon.
                \item Maroon xor yellow.
                \item Yellow iff orange.
                \item[] 
            \end{enumerate}
            \endgroup
            Conclusion: If maroon then orange.
            \vspace{6mm}
        };
        \node[box, right=0.01\textwidth of box4] (box5) {
            \textbf{Problem 5:}\\
            \vspace{1mm}
            \begingroup
            \setlist[enumerate]{itemsep=0pt, parsep=0pt, partopsep=0pt, topsep=0pt, leftmargin=*} % Removes space between list items and adjusts list margins
            Statements:\\
            \begin{enumerate}
                \item Purple iff yellow.
                \item Yellow iff blue.
                \item Blue xor orange.
                \item[] 
            \end{enumerate}
            \endgroup
            Conclusion: Purple xor orange.
            \vspace{6mm}
        };
        \node[box, right=0.01\textwidth of box5] (box6) {
            \textbf{Problem 6:}\\
            \vspace{1mm}
            \begingroup
            \setlist[enumerate]{itemsep=0pt, parsep=0pt, partopsep=0pt, topsep=0pt, leftmargin=*} % Removes space between list items and adjusts list margins
            Statements:\\
            \begin{enumerate}
                \item Gray iff yellow.
                \item Yellow xor olive.
                \item Olive iff black.
                \item[] 
            \end{enumerate}
            \endgroup
            Conclusion: If yellow then black.
            \vspace{6mm}
        };

        % Third row of beige boxes
        \node[box, below=0.01\textwidth of box4] (box7) {
            \textbf{Problem 7:}\\
            \vspace{1mm}
            \begingroup
            \setlist[enumerate]{itemsep=0pt, parsep=0pt, partopsep=0pt, topsep=0pt, leftmargin=*} % Removes space between list items and adjusts list margins
            Statements:\\
            \begin{enumerate}
                \item Blue iff red.
                \item Red xor white.
                \item White iff pink.
                \item[] 
            \end{enumerate}
            \endgroup
            Conclusion: If not blue then pink.
            \vspace{6mm}
        };
        \node[box, right=0.01\textwidth of box7] (box8) {
            \textbf{Problem 8:}\\
            \vspace{1mm}
            \begingroup
            \setlist[enumerate]{itemsep=0pt, parsep=0pt, partopsep=0pt, topsep=0pt, leftmargin=*} % Removes space between list items and adjusts list margins
            Statements:\\
            \begin{enumerate}
                \item Olive xor brown.
                \item Brown iff gray.
                \item Gray xor maroon.
                \item[] 
            \end{enumerate}
            \endgroup
            Conclusion: If not olive then maroon.
            \vspace{6mm}
        };
        \node[box, right=0.01\textwidth of box8] (box9) {
            \textbf{Problem 9:}\\
            \vspace{1mm}
            \begingroup
            \setlist[enumerate]{itemsep=0pt, parsep=0pt, partopsep=0pt, topsep=0pt, leftmargin=*} % Removes space between list items and adjusts list margins
            Statements:\\
            \begin{enumerate}
                \item Purple iff blue.
                \item Blue iff olive.
                \item Olive xor red.
                \item Red xor green.
                \item[] 
            \end{enumerate}
            \endgroup
            Conclusion: If purple then green.
            \vspace{3mm}
        };

        % Fourth row of beige boxes
        \node[box, below=0.01\textwidth of box7] (box10) {
            \textbf{Problem 10:}\\
            \vspace{1mm}
            \begingroup
            \setlist[enumerate]{itemsep=0pt, parsep=0pt, partopsep=0pt, topsep=0pt, leftmargin=*} % Removes space between list items and adjusts list margins
            Statements:\\
            \begin{enumerate}
                \item Brown iff yellow.
                \item Yellow xor green.
                \item Green iff purple.
                \item Purple iff olive.
                \item[] 
            \end{enumerate}
            \endgroup
            Conclusion: If brown then olive.
            \vspace{3mm}
        };
        \node[box, right=0.01\textwidth of box10] (box11) {
            \textbf{Problem 11:}\\
            \vspace{1mm}
            \begingroup
            \setlist[enumerate]{itemsep=0pt, parsep=0pt, partopsep=0pt, topsep=0pt, leftmargin=*} % Removes space between list items and adjusts list margins
            Statements:\\
            \begin{enumerate}
                \item Red iff maroon.
                \item Green xor olive.
                \item Maroon iff green.
                \item Olive xor brown.
                \item[] 
            \end{enumerate}
            \endgroup
            Conclusion: If red then brown.
            \vspace{3mm}
        };
        \node[box, right=0.01\textwidth of box11] (box12) {
            \textbf{Problem 12:}\\
            \vspace{1mm}
            \begingroup
            \setlist[enumerate]{itemsep=0pt, parsep=0pt, partopsep=0pt, topsep=0pt, leftmargin=*} % Removes space between list items and adjusts list margins
            Statements:\\
            \begin{enumerate}
                \item Blue iff brown.
                \item White iff green.
                \item Brown xor white.
                \item Green iff purple.
                \item[] 
            \end{enumerate}
            \endgroup
            Conclusion: If blue then purple.
            \vspace{3mm}
        };
    \end{tikzpicture}
  \caption{The task prompt (upper yellow box) as well as statements and conclusion for each propositional logic problem (lower gray boxes). In the task prompt, the placeholder \textit{``<statements and conclusion from below>''} is replaced with the actual statements and conclusion relevant to each problem. To enhance readability, we employ abbreviations within the problem statements. In the actual prompt, ``colorA iff colorB'' is replaced by ``There is a colorA marble in the box if and only if there is a colorB marble in the box''. Similarly, ``colorA xor colorB'' denotes ``Either there is a colorA marble in the box or else there is a colorB marble in the box, but not both''. Lastly, ``If colorA then colorB'' stands for ``If there is a colorA marble in the box then there is a colorB marble in the box''.}
  \label{fig:appendix_a_task_prompt}
\end{figure*}

\begin{table*}[htbp]
\centering
\renewcommand{\arraystretch}{1.2}
{\fontsize{8}{10}\selectfont
\begin{tabular}
{|>{\centering\arraybackslash}m{2.4cm}| *{7}{>{\centering\arraybackslash}m{1.4cm}|}}
\hline
  & \cellcolor{green!20} Supposition Following & \cellcolor{yellow!20} Chain Construction & \cellcolor{orange!20} Compound Conclusion & \cellcolor{mylightblue} Concatenation Strategy & \cellcolor{blue!20} Symbolic Strategy & Correct Answer & Sound Reasoning \\ \hline
\hline
Zephyr-7B-$\beta$ & $1.0$ & $0.94$ & $1.0$ & $1.0$ & $1.0$ & $1.0$ & $1.0$  \\
\hline
Mistral-7B-Instruct & $1.0$ & $0.9$ & $1.0$ & $1.0$ & $1.0$ & $1.0$ & $1.0$  \\
\hline
\LLaMASmall{} & $0.89$ & $0.95$ & $1.0$ & $0.79$ & $1.0$ & $1.0$ & $1.0$ \\
\hline
\LLaMAMedium{} & $0.88$ & $1.0$ & $0.85$ & $1.0$ & $1.0$ & $1.0$ & $1.0$ \\
\hline
\LLaMABig{} & $0.97$ & $1.0$ & $1.0$ & $1.0$ & $1.0$ & $1.0$ & $1.0$\\
\hline
\end{tabular}
}
\caption{Cohen's Kappa values to assess the inter-annotator agreement across different models and label categories.}
\label{tab:cohens_kappa}
\end{table*}

\subsection{Task Prompts}\label{subsec:appendix_a_task_prompts}
Figure \ref{fig:appendix_a_task_prompt} displays the task prompt and problem formulations employed in assessing the language models described in Section \ref{sec:Method}. Note that the prompt template, i.e. special tokens and their arrangements, might vary depending on the specific language model used. Within the task prompt (provided in the upper box), the problem statements and conclusion for a given problem are replaced with the corresponding problem formulations found in the lower gray boxes. In the final version of the prompt, the phrase ``colorA iff colorB'' is expanded to ``There is a colorA marble in the box if and only if there is a colorB marble in the box''. Similarly, ``colorA xor colorB'' is interpreted as ``There is either a colorA marble or a colorB marble in the box, but not both'', and ``If colorA then colorB'' is articulated as ``If there is a colorA marble in the box, then there is a colorB marble in the box''.

\subsection{Annotator Instructions}\label{subsec:appendix_a_annotator_instructions}
Our assessment of model responses involves a comprehensive independent review by two students who are specialized in the field of natural language processing and have expertise in manual data annotation. To ensure a high quality of annotations, we offer comprehensive training to both annotators. This training includes detailed explanations and extensive examples of the strategies identified by \citet{van_der_henst_strategies_2002}, complemented by a session dedicated to clarifying any questions that may emerge. Subsequently, the annotators are tasked with independently annotating practice examples, which serves to highlight and address any ambiguities in the annotation process. Only when both annotators are confident in their understanding of each strategy do we proceed. We instruct both annotators to independently go through each model response and mark parts where they identify a certain strategy to be employed. Each strategy is marked in a unique color code, which is afterwards converted into labels that signify the use of a particular strategy. In addition, we instruct both annotators to label whether the reasoning is sound, and the final conclusion of the model is correct. Furthermore, we ask them to classify any logical errors identified within the reasoning process. To maintain a high standard of annotation quality, annotators are instructed to review the model responses twice.

\subsection{Inter-Annotator Agreement}\label{subsec:appendix_a_inter_annotator_agreement}
To assess the reliability of our manual evaluation process (see Section \ref{sec:Method}), we quantify the inter-annotator agreement by calculating Cohen's Kappa for each category and model, as illustrated in Table \ref{tab:cohens_kappa}. Generally, the results indicate an almost perfect level of agreement across all categories and models, with Cohen's Kappa values ranging from $0.81 \leq \kappa \leq 1.0$. An exception is observed in the case of the \textit{concatenation strategy} applied by \LLaMASmall, for which we report a substantial agreement level, with a Kappa value of $ \kappa = 0.79$, slightly below the threshold for almost perfect agreement.

\subsection{Model Details}\label{subsec:appendix_a_model_details}
We report further details about the models used in this study in Table \ref{tab:llm_comparison}. In particular, we provide information about the number of parameters, context length and fine-tuning procedure for each model.

% Relative occurences of strategies found in the models' reasoning
\begin{table*}[htp]
\centering
\renewcommand{\arraystretch}{1.2}
{\fontsize{8}{10}\selectfont
\begin{tabular}
{>{\centering\arraybackslash}m{2.4cm} *{5}{>{\centering\arraybackslash}m{1.8cm}}}
\toprule
 Model & Base Model & Parameters & Context Length & Tokens & Fine-tuning \\ \hline
\midrule
Zephyr-7B-$\beta$ & Mistral & 7B & 8192 tokens & - & dSFT, AIF \\
\cmidrule{1-6}
Mistral-7B-Instruct & Mistral & 7B & 8192 tokens & - & SFT \\
\cmidrule{1-6}
LLaMA-2-7B-Chat & LLaMA-2 & 7B  & 4K tokens & 2.0T & SFT, RLHF \\
LLaMA-2-13B-Chat & LLaMA-2 & 13B  & 4K tokens & 2.0T & SFT, RLHF \\
LLaMA-2-70B-Chat & LLaMA-2 & 70B  & 4K tokens & 2.0T & SFT, RLHF \\
\bottomrule
\end{tabular}
}
\caption{Properties of the models used in this study. The context length refers to the base model's training. Tokens relate to the number of tokens in the pre-training data only. We use the following abbreviations for the fine-tuning procedure: supervised fine-tuning (SFT), reinforcement learning with human feedback (RLHF), distilled supervised fine-tuning (dSFT), and AI feedback through preferences (AIF). Information about the Llama 2 family is taken from \citet{touvron_llama_2023}, specifications for Mistral-7B-Instruct are provided by \citet{jiang_mistral_2023}. For Zephyr-7B-$\beta$, we consider the work of \citet{tunstall_zephyr_2023}. Dashes represent cases in which we could not find the respective information.}
\label{tab:llm_comparison}
\end{table*}

\section{Additional Quantitative Results}\label{sec:appendix_b_additional_quantitative_results}
In this segment, we present supplementary findings from our quantitative evaluation. Table \ref{tab:inf_strategies_frequencies_with_stdev} illustrates the frequencies with which the different language models employ inferential strategies when navigating the problems of propositional logic, as outlined in Section \ref{sec:Method}. Values denote percentages averaged across five distinct random seeds, accompanied by their standard deviation. Furthermore, we detail the proportions of correct final conclusions and sound reasoning. Note that all percentages are calculated relative to the overall count of tasks within the experimental framework.

\section{Annotated Model Responses}\label{sec:appendix_c_annotated_model_responses}
Within this section, we showcase examples of model responses that exemplify each inferential strategy identified in our study, as depicted in figures \ref{fig:appendix_c_supposition_following_llama2_70B_problem7}-\ref{fig:appendix_c_chain_symbolic_strategy_zephyr_7B_problem6}. Each figure is organized with the problem statement at the top, the model's response on the lower left, and the annotators' comments to the lower right. For an extensive array of model responses and annotations, we invite readers to explore our data repository at: \href{https://huggingface.co/datasets/mainlp/inferential\_strategies}{huggingface.co/datasets/mainlp/inferential\_strategies}.

% Relative occurences of strategies found in the models' reasoning
\begin{table*}[htpb]
\centering
\renewcommand{\arraystretch}{1.2}
{\fontsize{8}{10}\selectfont
\begin{tabular}
{>{\centering\arraybackslash}m{2.4cm} *{5}{>{\centering\arraybackslash}m{1.4cm}} !{\vrule width 1pt} *{2}{>{\centering\arraybackslash}m{1.4cm}}}
\toprule
 Model & \cellcolor{green!20} Supposition Following & \cellcolor{yellow!20} Chain Construction & \cellcolor{orange!20} Compound Conclusion & \cellcolor{mylightblue} Concatenation Strategy & \cellcolor{blue!20} Symbolic Strategy & Correct Answer & Sound Reasoning \\ \hline
\midrule
Zephyr-7B-$\beta$ & $60.0 \pm 12.2$ & $18.3 \pm 6.2\phantom{0}$ & $10.0 \pm 6.2\phantom{0}$ & $\phantom{0}1.7 \pm 3.3\phantom{0}$ & $20.0 \pm 11.3$ & $45.0 \pm 15.5$ & $25.0 \pm 10.5$  \\
\cmidrule{1-8}
Mistral-7B-Instruct & $35.0 \pm 6.2\phantom{0}$ & $10.0 \pm 3.3\phantom{0}$ & $35.0 \pm 9.7\phantom{0}$ & $\phantom{0}3.3 \pm 4.1\phantom{0}$ & $\phantom{0}8.3 \pm 7.5\phantom{0}$ & $55.0 \pm 10.0$ & $25.0 \pm 7.5\phantom{0}$  \\
\cmidrule{1-8}
\LLaMASmall{} & $20.0 \pm 6.7\phantom{0}$ & $20.0 \pm 15.5$ & $\phantom{0}6.7 \pm 3.3\phantom{0}$ & $\phantom{0}3.3 \pm 4.1\phantom{0}$ & $\phantom{0}1.7 \pm 3.3\phantom{0}$ & $46.7 \pm 6.7\phantom{0}$ & $\phantom{0}0.0 \pm 0.0\phantom{0}$ \\
\LLaMAMedium{} & $28.3 \pm 10.0$ & $36.7 \pm 12.5$ & $\phantom{0}6.7 \pm 3.3\phantom{0}$ & $\phantom{0}6.7 \pm 6.2\phantom{0}$ & $\phantom{0}0.0 \pm 0.0\phantom{0}$ & $40.0 \pm 8.2\phantom{0}$ & $15.0 \pm 6.2\phantom{0}$ \\
\LLaMABig{} & $45.0 \pm 8.5\phantom{0}$ & $50.0 \pm 7.5\phantom{0}$ & $\phantom{0}3.3 \pm 4.1\phantom{0}$ & $\phantom{0}1.7 \pm 3.3\phantom{0}$ & $\phantom{0}6.7 \pm 3.3\phantom{0}$ & $56.7 \pm 6.2\phantom{0}$ & $31.7 \pm 9.7\phantom{0}$\\
\bottomrule
\end{tabular}
}
\caption{Relative occurrences of inferential strategies employed by the different language models when solving the propositional problems. All values denote percentages averaged across 5 different random seeds with standard deviation. In addition, the percentages of correct final answers and sound reasoning are reported.}
\label{tab:inf_strategies_frequencies_with_stdev}
\end{table*}

\subsection{Supposition Following}\label{sec:appendix_c_supposition_following}
Figures from \ref{fig:appendix_c_supposition_following_llama2_70B_problem7} to \ref{fig:appendix_c_supposition_following_mistral_instruct_problem9} demonstrate the application of \textit{supposition following} by various models. For instance, Figure \ref{fig:appendix_c_supposition_following_llama2_70B_problem7} presents \LLaMABig's approach to problem 7, where the model supposes the absence of a blue marble in the box and logically infers the implications of this assumption to reach the valid conclusion. On the other hand, Figure \ref{fig:appendix_c_supposition_following_mistral_instruct_problem7} depicts Mistral-7B-Instruct's response to the same problem, where the model considers various combinations of marble in the box, drawing immediate conclusions that follow from the premises at hand. However, it does not explore deeper ramifications of these suppositions, thereby failing to deduce the validity of the conclusion. This showcases a common behavior we observe in models that employ \textit{supposition following} unsuccessfully. In Figure \ref{fig:appendix_c_supposition_following_mistral_instruct_problem9} the model approaches problem 9 by assuming the presence of an olive marble in the box, yet inferring disjointed intermediate conclusions that do not aid in solving the problem, thus failing to prove the logical validity of the problem.

\subsection{Chain Construction}\label{sec:appendix_c_chain_construction}
Figures \ref{fig:appendix_c_chain_construction_llama2_70B_problem9} to \ref{fig:appendix_c_chain_construction_llama2_70B_problem12} illustrate instances where models employ \textit{chain construction} to navigate the problems of propositional logic. In Figure \ref{fig:appendix_c_chain_construction_llama2_70B_problem9}, \LLaMABig{} adeptly forms a chain of conditional statements that bridge the antecedent of the conclusion to its consequent, effectively validating the conclusion's logical soundness. Conversely, Figure \ref{fig:appendix_c_chain_construction_llama2_70B_problem7} depicts a logical chain in which \LLaMABig{} erroneously concludes the nonexistence of a white marble based on the absence of a red marble, despite an exclusive disjunction linking the two. Despite this logical misstep, the model's final conclusion remains accurate, highlighting the discrepancy between the model's final answer and the soundness of its reasoning. In Figure \ref{fig:appendix_c_chain_construction_llama2_13B_problem10}, \LLaMAMedium{} constructs a chain correctly linking the antecedent of the final conclusion to its consequent. Nonetheless, it overlooks the negation present in one of the conditionals, resulting in a compromised reasoning chain. Figure \ref{fig:appendix_c_chain_construction_llama2_70B_problem5} presents a scenario where the model incorrectly attempts to validate an exclusive disjunction solely through a singular conditional sequence, a reasoning error not uncommon among human reasoners \cite{van_der_henst_strategies_2002}. Lastly, Figure \ref{fig:appendix_c_chain_construction_llama2_70B_problem12} highlights \LLaMABig's engagement in the inverse fallacy, inferring $\neg$ W $\rightarrow \neg$ G from the conditional W $\rightarrow$ G, mirroring a logical misjudgment frequently observed in human reasoning processes.

\subsection{Compound Strategy}\label{sec:appendix_c_compound_strategy}
The \textit{compound strategy} is illustrated in Figures \ref{fig:appendix_c_compound_strategy_mistral_instruct_problem9} to \ref{fig:appendix_c_compound_supp_strategy_mistral_instruct_problem8}. Figure \ref{fig:appendix_c_compound_strategy_mistral_instruct_problem9} presents Mistral-7B-Instruct's approach to problem 9, where it infers a biconditional relationship between the purple and olive marble from the first two premises. On the other hand, Figure \ref{fig:appendix_c_compound_strategy_llama2_70B_problem9} shows \LLaMABig's response to the same problem, formulating a sequence of compound inferences beyond the initial biconditional deduction, culminating in the correct final answer. Additionally, Figure \ref{fig:appendix_c_compound_supp_strategy_mistral_instruct_problem8} illustrates Mistral-7B-Instruct's approach to problem 8, in which the model initially generates compound conclusions derived from the problem statements, followed by \textit{supposition following} to explore the implications that the absence of an olive marble might have. However, despite the model's sound reasoning, its final answer is incorrect.

\subsection{Concatenation Strategy}\label{sec:appendix_c_concatenation_strategy}
Figure \ref{fig:appendix_c_concatenation_strategy_mistral_instruct_problem6} demonstrates the \textit{concatenation strategy}, where Mistral-7B-Instruct concatenates two intermediate deductions to form a single statement. It then uses the concatenated statement to infer the invalidity of the conclusion.

\subsection{Symbolic Strategy}\label{sec:appendix_c_symbolic_strategy}
The \textit{symbolic strategy} is exemplified in Figure \ref{fig:appendix_c_symbolic_strategy_llama2_70B_problem3}, where \LLaMABig{} employs a truth table to assess the conclusion's validity, albeit with errors leading to an incorrect result. Conversely, Figure \ref{fig:appendix_c_chain_symbolic_strategy_zephyr_7B_problem6} shows Mistral-7B-Instruct's application of \textit{chain construction} followed by the \textit{symbolic strategy}. The model makes false inferences while employing \textit{chain construction}, and further errs in its validation through logical calculus.

% === Qualitative examples ===

% Supposition Following
\begin{figure*}[htbp]
  \centering
  \input{tikz_files/appendix/C-Qualitative-Examples/Supposition-Following/problem7_llama2_70B_sup_following}
  \caption{The response (lower left box) of \LLaMABig{} to problem 7 (top box) of the problem set, illustrating \suppositionstrat~\textbf{supposition following}. After reformulating the statements of the problem at hand, the model starts to reason about the problem by assuming the absence of a blue marble in the box. Subsequently, it traces the consequences of that supposition, drawing intermediate conclusions about the presence or absence of certain marbles, until it formulates a final conclusion. In this example, the model correctly reasons about the validity of the conclusion.}
  \label{fig:appendix_c_supposition_following_llama2_70B_problem7}
\end{figure*}

\begin{figure*}[htbp]
  \centering
  \input{tikz_files/appendix/C-Qualitative-Examples/Supposition-Following/problem7_mistral_instruct_sup_following}
  \caption{An exemplary model response of  Mistral-7B-Instruct (lower left box) to problem 7 (top box) illustrating \suppositionstrat~\textbf{supposition following}. The model successively assumes combinations of marbles in the box, and infers the immediate consequences from the premises provided. However, it does not extend its reasoning beyond the direct outcomes of each supposition, thereby failing to deduce the validity of the conclusion.}
  \label{fig:appendix_c_supposition_following_mistral_instruct_problem7}
\end{figure*}

\begin{figure*}[htbp]
  \centering
  \input{tikz_files/appendix/C-Qualitative-Examples/Supposition-Following/problem9_mistral_instruct_sup_following}
  \caption{An exemplary model response of  Mistral-7B-Instruct (lower left box) to problem 9 (top box) illustrating \suppositionstrat~\textbf{supposition following}. The model supposes the presence of an olive marble in the box and traces the consequences of that supposition. However, it derives disjointed intermediate conclusions that do not aid in solving the problem, failing to solve the task at hand.}
  \label{fig:appendix_c_supposition_following_mistral_instruct_problem9}
\end{figure*}

% Chain Construction
\begin{figure*}[htbp]
  \centering
  \input{tikz_files/appendix/C-Qualitative-Examples/Chain-Construction/problem9_llama2_70B_chain_construction}
  \caption{The response (lower left box) of \LLaMABig{} to problem 9 (top box) of the problem set, illustrating \chainstrat~\textbf{chain construction}. The model correctly constructs a chain of conditionals leading from the antecedent of the final conclusion to its consequent.}
  \label{fig:appendix_c_chain_construction_llama2_70B_problem9}
\end{figure*}

\begin{figure*}[htbp]
  \centering
  \input{tikz_files/appendix/C-Qualitative-Examples/Chain-Construction/problem7_llama2_70B_chain_construction}
  \caption{The response (lower left box) of \LLaMABig{} to problem 7 (top box) of the problem set, illustrating \chainstrat~\textbf{chain construction}. The model constructs a chain of conditionals leading from the antecedent of the final conclusion to its consequent. However, it fails to understand the implication of the exclusive disjunction in the second statement of the problem description, leading to a faulty reasoning trace. Despite its invalid reasoning, the model's final answer is correct.}
  \label{fig:appendix_c_chain_construction_llama2_70B_problem7}
\end{figure*}

\begin{figure*}[htbp]
  \centering
  \input{tikz_files/appendix/C-Qualitative-Examples/Chain-Construction/problem10_llama_13B_chain_construction}
  \caption{The response (lower left box) of \LLaMAMedium{} to problem 10 (top box) of the problem set, illustrating \chainstrat~\textbf{chain construction}. The model constructs a chain of conditionals leading from the antecedent of the final conclusion to its consequent. However, it fails to account for the negation of the second conditional's consequent, leading to a faulty reasoning trace.}
  \label{fig:appendix_c_chain_construction_llama2_13B_problem10}
\end{figure*}

\begin{figure*}[htbp]
  \centering
  \input{tikz_files/appendix/C-Qualitative-Examples/Chain-Construction/problem5_llama2_70B_chain_construction}
  \caption{The response (lower left box) of \LLaMABig{} to problem 5 (top box) of the problem set, illustrating \chainstrat~\textbf{chain construction}. The model constructs a chain of conditionals proving one case of the exclusive disjunction. However, it fails to account for the other conditional case, i.e. $\neg$ P $\rightarrow$ O, therefore failing to prove the logical validity of the conclusion.}
  \label{fig:appendix_c_chain_construction_llama2_70B_problem5}
\end{figure*}

\begin{figure*}[htbp]
  \centering
  \input{tikz_files/appendix/C-Qualitative-Examples/Chain-Construction/problem12_llama2_70B_chain_construction}
  \caption{The response (lower left box) of \LLaMABig{} to problem 12 (top box) of the problem set, illustrating \chainstrat~\textbf{chain construction}. The model constructs a chain of conditionals leading from the antecedent of the final conclusion to its consequent. However, it makes a series of mistakes when constructing the chain of conditionals. For instance, it infers the absence of the green marble by denying the presence of the white marble, i.e. Blue $\rightarrow \neg$ W; W $\rightarrow$ G; therefore Blue $\rightarrow \neg$ G by assuming that $\neg$ W $\rightarrow \neg$ G, which is a common logical error known as the \textit{fallacy of the inverse}.}
  \label{fig:appendix_c_chain_construction_llama2_70B_problem12}
\end{figure*}

% Compound Strategy
\begin{figure*}[htbp]
  \centering
  \input{tikz_files/appendix/C-Qualitative-Examples/Compound-Strategy/problem9_mistral_instruct_comp_strat}
  \caption{The response (lower left box) of Mistral-7B-Instruct to problem 9 (top box) of the problem set, illustrating the \compoundstrat~\textbf{compound strategy}. Based on the first two premises of the problem description, the model draws a compound conclusion, establishing equivalence between the purple and olive marble in the box. However, Mistral-7B-Instruct fails to draw additional intermediate conclusions that would be required to deduce the logical validity of the conclusion in the problem statement.}
  \label{fig:appendix_c_compound_strategy_mistral_instruct_problem9}
\end{figure*}

\begin{figure*}[htbp]
  \centering
  \input{tikz_files/appendix/C-Qualitative-Examples/Compound-Strategy/problem9_llama2_70B_comp_strat}
  \caption{The response (lower left box) of \LLaMABig{} to problem 9 (top box) of the problem set, illustrating the \compoundstrat~\textbf{compound strategy}. The model draws a series of compound conclusions to deduce the logical validity of the conclusion in the problem statement.}
  \label{fig:appendix_c_compound_strategy_llama2_70B_problem9}
\end{figure*}

\begin{figure*}[htbp]
  \centering
  \input{tikz_files/appendix/C-Qualitative-Examples/Compound-Strategy/problem8_mistral_instruct_comp_supp_strat}
  \caption{The response (lower left box) of Mistral-7B-Instruct to problem 8 (top box) of the problem set, illustrating the \compoundstrat~\textbf{compound strategy} and \suppositionstrat~\textbf{supposition following}. Based on the first two premises of the problem description, the model first draws a compound conclusion, establishing that a gray marble follows from the absence of an olive marble. Subsequently, it uses this intermediate conclusion, together with the third premise, to draw another compound conclusion about the absence of the maroon marble. The model then switches to \textit{supposition following}, tracing the consequences of the absence of the olive marble, inferring the final conclusion that there cannot be a maroon marble. However, despite the model's correct reasoning, it deduces the wrong answer: "True".}
  \label{fig:appendix_c_compound_supp_strategy_mistral_instruct_problem8}
\end{figure*}

% Concatenation Strategy
\begin{figure*}[htbp]
  \centering
  \input{tikz_files/appendix/C-Qualitative-Examples/Concatenation-Strategy/problem6_mistral_instruct_concat_strat}
  \caption{The response (lower left box) of Mistral-7B-Instruct to problem 6 (top box) of the problem set, illustrating the \concatstrat~\textbf{concatenation strategy}. Mistral-7B-Instruct concatenates the intermediate conditional conclusion (Y $\rightarrow \neg$ O) and the third premise of the problem statement (O $\leftrightarrow$ B) to form the concatenated conclusion Y $\rightarrow \neg$ (O $\leftrightarrow$ B). Based on that conclusion, the model infers that the conclusion in the problem statement does not logically follow from the premises at hand.}
  \label{fig:appendix_c_concatenation_strategy_mistral_instruct_problem6}
\end{figure*}

% Symbolic Strategy
\begin{figure*}[htbp]
  \centering
  \input{tikz_files/appendix/C-Qualitative-Examples/Symbolic-Strategy/problem3_llama2_70B_symbolic_strat}
  \caption{The response (lower left box) of \LLaMABig{} to problem 3 (top) of the problem set, illustrating the \symbolicstrat~\textbf{symbolic strategy}. The model constructs a truth table to infer the validity of the conclusion given in the problem statement. However, the model produces errors in the truth table, resulting in flawed reasoning.}
  \label{fig:appendix_c_symbolic_strategy_llama2_70B_problem3}
\end{figure*}

\begin{figure*}[htbp]
  \centering
  \input{tikz_files/appendix/C-Qualitative-Examples/Symbolic-Strategy/problem6_zephyr_7B_symbolic_strat}
  \caption{The response (lower left box) of Zephyr-7B-$\beta$ to problem 6 (top box) of the problem set, illustrating \chainstrat~\textbf{chain construction} and the \symbolicstrat~\textbf{symbolic strategy}. The model first constructs a chain of conditionals to prove the validity of the conclusion, linking relevant entities in premise two and three of the problem statement. Subsequently, the model \textit{``explains''} its reasoning by employing the symbolic strategy, converting statements into formal logic and operating on them. Note that the model makes several logical errors on its way to prove the logical validity of the final conclusion.}
  \label{fig:appendix_c_chain_symbolic_strategy_zephyr_7B_problem6}
\end{figure*}

\end{document}